# R-BERT-CNN: Drug-target interactions extraction from biomedical literature

BioCreative Challenge VII Track 1 submission

Jehad Aldahdooh[1,2*], Ziaurrehman Tanoli[1*], Jing Tang[1,3*]
1. Research Program in Systems Oncology, Faculty of medicine, University of Helsinki, Finland;
2. Doctoral Program in Computer Science, University of Helsinki, Finland;
3. Department of Mathematics and Statistics, University of Helsinki;   * Correspondence

*Abstract*—In this research, we present our work participation for the DrugProt task of BioCreative VII challenge. Drug-target interactions (DTIs) are critical for drug discovery and repurposing, which are often manually extracted from the experimental articles. There are >32M biomedical articles on PubMed and manually extracting DTIs from such a huge knowledge base is challenging. To solve this issue, we provide a solution for Track 1, which aims to extract 10 types of interactions between drug and protein entities. We applied an **Ensemble Classifier** model that combines BioMed-RoBERTa, a state of art language model, with Convolutional Neural Networks (CNN) to extract these relations. Despite the class imbalances in the BioCreative VII DrugProt test corpus, our model achieves a good performance compared to the average of other submissions in the challenge, with the micro F1 score of 55.67% (and 63% on BioCreative VI ChemProt test corpus). The results show the potential of deep learning in extracting various types of DTIs.

I. INTRODUCTION

During the past decade, overall efforts in cancer drug discovery have made a clear shift to focus on targeted drugs directed towards deregulated proteins in cancer tissues. There has been an extensive effort to sequence cancer genomes in large patient cohorts, sparking expectations to identify novel drug-targets for more effective and selective treatment opportunities (1). Hundreds of such targeted drugs entered clinical trials, but often have disappointing efficacies due to varying treatment responses. This is most likely because we do not have sufficient understanding on the comprehensive drug-target interactions (DTIs), and how these interactions are leading to treatment efficacy as well as adverse effects (2). As a result, our understanding on drug mechanisms of action is limited, even if the drugs are successfully tested in clinical trials. For example, a recent study has shown that many putative targets for clinical trial cancer drugs failed to explain their efficacy, suggesting that off-targets are more likely the major contributors to the clinical efficacy (3). Promiscuous on-target and off-target interactions are inherent attributes for many drug-like small molecules. To completely understand the drug mechanism of action, there is a clear need to leverage vast volumes of experimental articles to facilitate the modeling and identification of novel drug-targets.

There are dozens of databases that provide manually extracted DTIs, among these are ChEMBL (4), BindingDB (5), PubChem (6), DrugBank (7), DrugCentral (8), GtopDB (9) and DGiDB (10). However, none of these drug-target databases provide target coverage for approved drugs at the whole proteome level, and only 11% of the human proteome are currently targeted by small molecules (2). To build a consensus knowledge base for drug–target interactions, we have recently introduced DrugTargetCommons, a community-based crowdsourcing initiative to improve the curation and wider use of drug–target bioactivity data for drug discovery, target identification, and repurposing applications (11,12). The web-based DrugTargetCommons platform aims to standardize the curation and annotation of heterogeneous compound-target bioactivity measurements and provide one of the most comprehensive, reproducible, and sustainable bioactivity knowledge base for end users (13).

However, manually curating the articles and using brute force to extract drug–target interaction data are daunting tasks. Therefore, we aim to develop text-mining based approaches to build a central DTIs repository that contains major clinical and pre-clinical drugs along with experimentally tested protein targets. Specifically, it will extract and annotate the bioactivity values and experimental protocols (such as assay and detection technology) that characterize the DTIs. More comprehensive drug-target profiles can be derived as high-quality drug-target bioactivity data is collected, allowing us to develop machine learning models for predicting DTIs. Experimental approaches can be used to further validate potentially predicted DTIs. Recently we have introduced BERT to identify potential articles from PubMed (14). The next step is to further extract the detailed drug-target information from the literature.

The BioCreative (Critical Assessment of Information Extraction systems in Biology) challenges provide a community-wide effort for evaluating text mining and information extraction systems applied to the biological domain. Both BioCreative VII track 1 (https://biocreative.bioinformatics.udel.edu/tasks/biocreative-vii/track-1/) and BioCreative VI track 5 (http://www.biocreative.org/tasks/biocreative-vi/track-5/) challenges aim to promote the development and evaluation of systems that can automatically detect and classify the relations between compounds and proteins (15, 25). In the ChemProt track, the organizers manually annotate a chemical–protein relation corpus composed of 2,432 PubMed abstracts, divided into a training set (1,020 abstracts), development set (612 abstracts) and test set (800 abstracts) (15). The task of ChemProt includes only five types of chemical–protein relations, the biological properties upregulator (CPR: 3), downregulator (CPR: 4), agonist (CPR: 5), antagonist (CPR: 6) and substrate (CPR: 9). Following the ChemProt track, the organizers generated a new manually annotated corpus called DrugProt corpus (25) which considers 13 types of interactions described in Table I.

By participating in the DrugProt BioCreative challenge, we aim to obtain hands-on experience on the most comprehensive training dataset and test its potential for developing deep learning based methods. We will apply the successfully trained models on the whole PubMed to extract a comprehensive collection of novel DTIs.

## II. TASK AND DATA

Track description Track 1 of the challenge involves PubMed abstract level relation extraction of chemical-protein relations. In this track, a manually annotated corpus called DrugProt is provided for training, development and testing purposes. In training and validation datasets, three main files are provided: a) pmids and abstracts texts, b) all chemical and gene mentions, c) all binary relationships between entities with pre-defined 13 biological relation types, such as agonists, antagonists, inhibitors, and regulators. Agonists activate proteins to produce the desired response, whereas antagonists prevent the activation. Inhibitor is a substance that slows or prevents a chemical reaction. Regulators are associated more with metabolism. The training data obtained using DrugProt corpus consisted of 46,274 chemicals and 43,255 proteins reported in 3,500 articles. The statistics of the DrugProt dataset are described in Table I. In this study, we aim to identify 10 types of interactions between chemical and biological molecules.

TABLE I.  STATISTICS OF DRUGPROT DATASET

|  | Train | Dev-set | Test |
|---|---|---|---|
| Document | 3,500 | 750 | 10,750 |
| Chemical | 46,274 | 9,853 | 143,767 |
| Protein | 43,255 | 9,005 | 167,038 |
| Positive relations | 17,288 | 3,765 | - |
| ANTAGONIST | 972 | 218 | - |
| INHIBITOR | 5,392 | 1,150 | - |
| AGONIST | 658 | 131 | - |
| ACTIVATOR | 1428 | 246 | - |
| INDIRECT-UPREGULATOR | 1378 | 302 | - |
| INDIRECT-DOWNREGULATOR | 1,330 | 332 | - |
| PART-OF | 886 | 258 | - |
| DIRECT-REGULATOR | 2,250 | 458 | - |
| SUBSTRATE | 2,003 | 495 | - |
| PRODUCT-OF | 921 | 158 | - |
| AGONIST ACTIVATOR | 29 | 10 | - |
| AGONIST-INHIBITOR | 13 | 2 | - |
| SUBSTRATE_PRODUCT-OF | 25 | 3 | - |

Preprocessing the text is a crucial step in text mining before feeding the text to the model. One of the most common libraries for text preprocessing techniques is NLTK (Natural Language Toolkit) in python (28).

We have pre-processed the data as follows:

1. We replace the tab-delimiter between title and abstract with a single space to follow the entity offsets according to the guidelines.

2. We split the raw documents (title and abstract) into sentences. Since the cross-sentence relations are rare in the corpus, we only focus on the relation occurring in the same sentence.

3. We deal with some possible bad separations of the tokenizer, for instance sentences ending with 'vivo', and 'Vmax'. However, these characters are not the ending of the sentences.

4. We assign special tags to the chemicals ($chemical entity name$) and proteins (#protein entity name#) regardless of their positions. We have not replaced the entity names with unified chemical or gene words in the sentence.

5. There are many entities in the sentences without any relationship. We assigned a new relation called 'Other' for those cases.

6. We decided to delete three less represented interaction types including AGONIST-ACTIVATOR, AGONIST-INHIBITOR, and SUBSTRATE_PRODUCT-OF.

For the first submission (official scores according to the deadline), we just use the DrugProt dataset for training, validation and testing the Model I (see details in section III). We used all the 13 interaction types in the training datasets and ignored imbalance class distributions, which has greatly affected the micro F1 score. However, while training Model II (see details in section III), we ignored the three least represented interaction types i.e AGONIST-ACTIVATOR, AGONIST-INHIBITOR and SUBSTRATE_PRODUCT-OF.

Each of these interaction types has less than 30 samples, which were not enough to sufficiently train the model.

For training the model II (results submitted after the deadline and before paper submission), we combined ChemProt and DrugProt datasets. ChemProt test data is used as a hidden golden test data to evaluate our system by using the DrugProt (BioCreative VII) evaluation library to generate the performance metrics.

### III. METHODS

Our model is an extension of BioMed-RoBERTa-base (16) to extract the relations (interaction types). BioMed-RoBERTa-base is a language model built based on the RoBERTa base model (17) with an additional pre-training of 12.5K steps. It was trained on more scientific papers (2.68M) from the Semantic Scholar. The training data contains 7.55B tokens and have both abstracts and full texts.

We adapt the BioMed-RoBERTa-base model for the relation classification task by encoding each relation pair of chemical protein mentions corresponding to the abstract text as in the following example format:

<s>This study therefore characterized the binding of $DF$ to the #sigma receptors# and NMDA - linked PCP sites and examined the anticonvulsant as well as locomotor effects of DF in mice in comparison with those of DM and DR.</s>

In the first model (Model I), we have only used the embedding vector of the BioRoBERTa [CLS] token from the last hidden layer as a representation of each textual sequence. It is further processed by two fully connected layers and a SoftMax activation function.

Our main model called R-BERT-CNN (Model II) shown in Fig. 1 has different parts. The first part is the BioMed-RoBERTa-base model in which the pre-processed input text sequence with length 512 passes through 12 layers. The second part inspired from (18) is CNN which utilizes the output of the last 4 layers from the BioMed-RoBERTa-base model. In the BERT paper (19) , it shows that the best performance could be achieved by concatenating the last 4 layers outputs to obtain the contextualized representations. The concatenated embedding layers are then passed to 48 convolutional filters with three different kernel sizes: 768*3, 768*4, 768*5 (each size with 16 filters). The output of each convolutional layer is fed to the ReLU activation function and a max-pooling layer. The results of the max-pooling layers are then concatenated and passed through a fully connected layer. The final part of the R-BERT-CNN model inspired from (20) is the vector representations for chemical protein target entities, produced by averaging the last hidden state vectors from BioMed-RoBERTa-base for each entity, which are passed to two different fully connected layers. Finally, all fully connected layers outputs are concatenated and fed to a final dense layer. Then, it is followed by a classification layer that computes the cross-entropy loss for classification with mutually exclusive classes.

In order to solve the problem of data imbalance, we used PyTorch sampler module to sample the data (26,27). The sampling method used is WeightedRandomSampler, which assigns the classes with different weights according to their total count in the dataset. Then, it selects the data according to the weight of each sample. Table II shows the hyper-parameters that are used to fine-tune the BioMed-RoBERTa-base model.

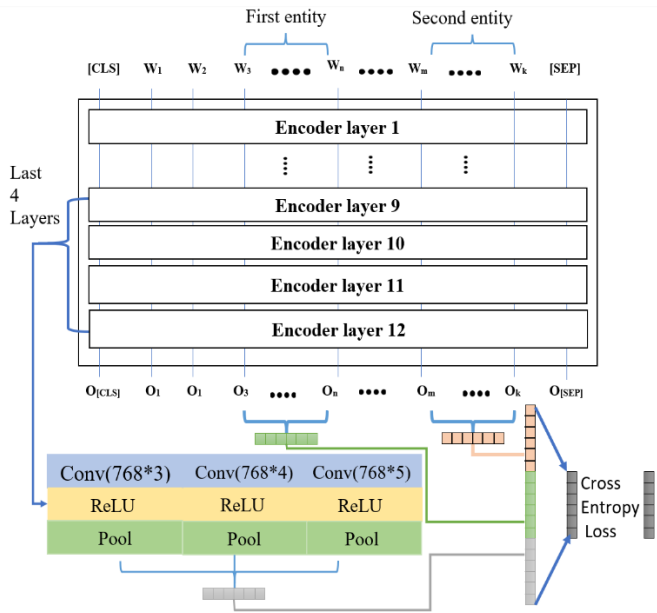

Fig. 1. Overall System Architecture of R-BERT-CNN.

TABLE II. MODEL II HYPER-PARAMETERS

| Dropout rate | 0.5 |
|---|---|
| Learning rate | 3e-05 |
| Epochs | 7 |
| Max sequence length | 512 |
| Batch size | 32 |
| Adam epsilon | 1e-8 |
| Gradient accumulation steps | 1 |
| Max grad norm | 1 |
| Weight decay | 0 |
| Num warmup steps | 0 |

### IV. RESULTS

The micro F1-score is used to assess the quality of multi-classification problems by aggregating the contribution of each class. The micro F1, micro recall and micro precision scores are defined using the following equations.

$$Micro\ F1 - score = \frac{2 \times Micro-precision \times Micro-precision}{Micro-precision + Micro-precision}$$

$$Micro\ precision = \frac{TP1 + TP2 \ldots TPn}{TP1 + TP2 + ..TPn + FP1 + FP2 + ..FPn}$$

$$Micro\ recall = \frac{TP1 + TP2 \ldots TPn}{TP1 + TP2 + ..TPn + FN1 + FN2 + ..FNn}$$

As mentioned in section III, model I was a simple trial to assess performance of a pre-trained model with an imbalance dataset (using DrugProt dataset). Since we include all the relations (interaction types) including even less represented ones, we expected the model to behave poorly (bad prediction results). The result for the model I on the DrugProt test dataset gives 6.3% F1 micro score.

TABLE III.  RESULTS OF MODEL II

| Relation | DrugProt test data | | | ChemProt test data | | |
|---|---|---|---|---|---|---|
| | P | R | F1 | P | R | F1 |
| ANTAGONIST | 0.59 | 0.98 | 0.73 | 0.61 | 0.96 | 0.75 |
| INHIBITOR | 0.61 | 0.80 | 0.69 | 0.67 | 0.88 | 0.76 |
| AGONIST | 0.51 | 0.91 | 0.66 | 0.40 | 1 | 0.57 |
| ACTIVATOR | 0.45 | 0.86 | 0.59 | 0.38 | 0.99 | 0.55 |
| INDIRECT-UPREGULATOR | 0.40 | 0.87 | 0.55 | 0.50 | 0.99 | 0.66 |
| INDIRECT-DOWNREGULATOR | 0.42 | 0.76 | 0.54 | 0.54 | 0.97 | 0.69 |
| PART-OF | 0.33 | 0.81 | 0.47 | 0.40 | 1.0 | 0.57 |
| DIRECT-REGULATOR | 0.35 | 0.69 | 0.47 | 0.42 | 0.94 | 0.58 |
| SUBSTRATE | 0.31 | 0.80 | 0.45 | 0.35 | 0.99 | 0.52 |
| PRODUCT-OF | 0.30 | 0.76 | 0.43 | 0.34 | 1.0 | 0.50 |
| Global results across all interactions types | 0.43 | 0.80 | 0.56 | 0.47 | 0.95 | 0.63 |

R-BERT-CNN model achieved better results over both DrugProt and ChemProt test datasets (Table 3) by skipping samples from the three least represented classes. Using 10 interaction types, we achieved the micro F1 score of 0.56 as shown in Table 3. The ANTAGONIST, INHIBITOR and AGONIST interaction types obtained relatively better individual performances with F1 scores: 0.73, 0.69 and 0.66 respectively. This could be due to the fact that the model was able to better understand these interaction types and see similar examples during the training process. Another reason could be the better data representation for these interaction types in terms of chemical and protein names in the actual text.

The model finds difficulty (F1 < 0.47) to capture the last four interaction types i.e. part-of, direct-regulator, substrate, product-of. We might need to obtain additional high quality samples for these interaction types to have better training of the model. These interaction types are difficult to curate/extract, and hence require significant manual curation efforts.

## V. DISCUSSION

Target coverage at full proteome scale is important to infer the mechanism of actions for approved drugs (21). However, many targeted drugs are annotated only with the primary putative targets, while off-target information remains poorly understood (22). To address this challenge, we need to develop efficient computational approaches to mine the existing datasets from publications.

During this challenge, we developed a R-BERT-CNN model to detect and classify the relations between chemicals and biological molecules. The results showed the micro F1 score of 56% across the 10 interaction types. In particular, the ANTAGONIST and INHIBITOR interaction types achieved the micro F1 score of approximately >68%. We will further implement error analysis on the false positive and false negative predictions (29), and seek to improve the models to predict drug–target interactions with higher accuracy by adding more manually curated ground truth data. For example, using preliminary analysis (14), we identified ~2.1M articles, which are likely to contain DTIs. We also extracted the associated chemical and protein entities. Models developed in this research, if applied on those ~2.1M articles dataset will pave the way to drug-target data collection, which will be used as a benchmark dataset to facilitate machine learning applications, explore structure–activity relationships, and identify predictive features for a ligand–receptor binding. In a pilot study, we evaluated multiple deep learning-based chemical fingerprints in terms of their prediction accuracy for drug combination sensitivity and synergy and found that Graph Isomorphism Network (Infomax) and Variational Autoencoder (VAE)-based fingerprints achieved the best performance (23). We will validate the novel targets (identified by the computational methods) using the experimental methods such as thermal proteome profiling (24). The resulting drug–target interaction networks can be integrated with chemical and pharmacological data to facilitate drug discovery and drug repurposing.

ACKNOWLEDGMENT

The work is supported by the European Research Council (DrugComb, No. 716063), the EU H2020 (EOSC-LIFE, No. 824087), and the Academy of Finland (No. 317680). The computational resources are provided by the Finnish IT Center for Science (CSC).